\documentclass[runningheads]{llncs}
\usepackage{graphicx}

\usepackage{tikz}
\usepackage{comment}
\usepackage{amsmath,amssymb} 
\usepackage{color}

\usepackage{caption}
\usepackage{subcaption}
\usepackage{graphicx}
\usepackage{amsmath}
\usepackage{amssymb}
\usepackage{booktabs}
\usepackage{mathtools}
\usepackage{algorithm2e}

\newcommand{\N}{\mathbb{N}}

\newcommand{\R}{\mathbb{R}}

\newcommand{\calV}{\mathcal{V}}

\newcommand{\norm}[1]{\left\Vert #1 \right\Vert}
\newcommand{\set}[1]{\left\lbrace #1\right\rbrace}

\DeclareMathOperator*{\argmax}{arg\,max}

\newcounter{RonCounter}

\newcounter{StefanCounter}

\newcounter{DACounter}

\newcounter{JoanCounter}

\newcommand{\ie}{\textit{i}.\textit{e}.}
\newcommand{\eg}{\textit{e}.\textit{g}.}

\usepackage[accsupp]{axessibility}  

\begin{document}
\pagestyle{headings}
\mainmatter
\def\ECCVSubNumber{100}  

\title{Cartoon Explanations of Image Classifiers} 

\titlerunning{Cartoon Explanations of Image Classifiers}
%
\author{Stefan Kolek\inst{1} \and
Duc Anh Nguyen\inst{1} \and
Ron Levie\inst{3} \and Joan Bruna \inst{4} \and Gitta Kutyniok \inst{1,2}}
\authorrunning{S. Kolek et al.}
%
\institute{Ludwig Maximilian University of Munich, Germany \email{\{kolek,danguyen,kutyniok\}@math.lmu.de} 
\and 
University of Tromsø, Norway
\and
Technion-Israel Institute of Technology, Israel\\
\email{levieron@technion.ac.il}\\
\and
New York University, NY, USA\\
\email{bruna@cims.nyu.edu}}
\maketitle

\begin{abstract}
We present \emph{CartoonX} (Cartoon Explanation), a novel model-agnostic explanation method tailored towards image classifiers and based on the rate-distortion explanation (RDE) framework. Natural images are roughly piece-wise smooth signals---also called cartoon-like images---and tend to be sparse in the wavelet domain. CartoonX is the first explanation method to exploit this by requiring its explanations to be sparse in the wavelet domain, thus extracting the \emph{relevant piece-wise smooth} part of an image instead of relevant pixel-sparse regions. We demonstrate that CartoonX can reveal novel valuable explanatory information, particularly for misclassifications. Moreover, we show that CartoonX achieves a lower distortion with fewer coefficients than state-of-the-art methods.
\end{abstract}


\section{Introduction}
\label{sec:intro}

Powerful machine learning models such as deep neural networks are inherently opaque, which has motivated numerous explanation methods over the last decade (see for example the survey by \cite{Das2020OpportunitiesAC}). A significant fraction of the research literature has focused on explaining image classifications due to both the practical relevance of computer vision tasks and the ease at which heatmaps can communicate explanatory information. Despite the great variety in methods and explanation philosophies, all current methods share the following characteristic: they operate in pixel space. Roughly speaking, existing explanation methods for image classifiers either allocate additive attribution scores to each (super)pixel or optimize a deletion mask on the pixel coefficients to mark a relevant set of pixels. The result is typically a pixel-sparse and jittery explanation. We challenge the conventional approach to explain in pixel space by successfully applying the rate-distortion explanation (RDE) framework \cite{RDE_original_2019,in_distribution_cosmas_2020} in the wavelet domain of images. Our novel explanation method, \emph{CartoonX}, extracts the relevant piece-wise smooth part of an image. Instead of demanding sparsity in pixel space, as in \cite{RDE_original_2019,chang_GAN_explanation_2018}, CartoonX demands sparsity in the wavelet domain, which produces piece-wise smooth explanations. Piece-wise smooth images are also known as \emph{cartoon-like images} \cite{kutyniok2010}---a class of 2D signals that has been well studied, and for which wavelets provides an efficient representation system \cite{Romberg06wavelet-domainapproximation}. 
Our work makes the following contributions.

\emph{Reformulation and reinterpretation of the RDE framework:} We reformulate the RDE framework in a more general manner with enhanced flexibility in the input representation to accommodate complex interpretation queries such as ``What is the piece-wise smooth part of the input signal that leads to its model decision?". Thereby, we reinterpret RDE as a simplification of the input signal, which is interpretable to humans and adheres to a meaningful interpretation query. The simplification is achieved by demanding sparsity in a suitable representation system, which sparsely represents the class of explanations that are desirable for the interpretation query.

\emph{CartoonX, a novel explanation method tailored to image classifiers:} CartoonX is the first explanation method to extract the relevant piece-wise smooth part of an image instead of relevant pixel sparse regions. This is achieved by demanding sparsity in the wavelet domain of images, where sparsity translates into piece-wise smooth images. We demonstrate that our piece-wise smooth explanations can reveal relevant piece-wise smooth patterns that are not easily visible with existing pixel-based methods. Quantitatively, we also corroborate that CartoonX achieves a lower distortion in the model output using fewer coefficients than other state-of-the-art methods.

\section{Related Work}
The Rate-Distortion Explanation (RDE) framework was first introduced in \cite{RDE_original_2019}, and extended in \cite{in_distribution_cosmas_2020}, as a mathematically well-founded and intuitive explanation framework. RDEs are model-agnostic explanations and inspired by rate-distortion theory, which studies lossy-data compression. An explanation in RDE consists of a relatively sparse mask over the input features, highlighting the relevant set of features. The mask is optimized to produce low distortion in the model output after applying perturbations to the unselected features in the input while remaining relatively sparse. The authors of \cite{in_distribution_cosmas_2020} also applied RDE to non-canonical input representations to explain model decisions in challenging domains such as audio classification  \cite{engel2017neural} and radio-map estimation \cite{levie2019radiounet,levieIEEE2020}.
The explanation principle of optimizing a mask $s\in [0,1]^n$ was first proposed by \cite{Fong_vedaldi_2017} who explained image classification decisions by considering one of the two ``deletion games": (1) optimizing for the smallest deletion mask that causes the class score to drop significantly or (2) optimizing for the largest deletion mask that has no significant effect on the class score.  The original RDE approach \cite{RDE_original_2019} is based on the second deletion game. We decided to work within the RDE framework, due to its flexible mathematical formulation. However, we note that other viable mask-based explanation frameworks such as RISE \cite{Petsiuk2018RISERI}, which does not assume access to the model gradient, exist.
Other explanation methods developed by the research community are typically either (1) gradient-based such as Smoothgrad \cite{Smooth_Grad_2017}, Integrated Gradients \cite{Integrated_gradient_2017_sundararajan}, and Grad-CAM \cite{Grad_CAM_2017}, (2) surrogate models such as LIME \cite{LIME_2016}, (3) based on propagation of activations in neurons such as LRP \cite{Layerwise_relevance_prop2015,DeepLIFT_2017}, and DeepLIFT \cite{DeepLIFT_2017}, (4) based on Shapely values from game-theory \cite{SHAP_neurips_2017}, (6) concept-based such as Concept Activation Vectors \cite{Kim2018InterpretabilityBF}, or (7) based on generative causal explanations \cite{generativecausal2020}. Also related are methods that were developed to explain individual neurons such as in \cite{preferred_inputs_for_neurons,dhamdhere2018how}.  To our knowledge, all existing explainability methods operate in pixel space and all methods looking for sparse explanations demand sparsity in pixel space \cite{RDE_original_2019,Fong_vedaldi_2017,chang_GAN_explanation_2018}.

\section{Background: RDE}

In this section, we review the rate-distortion explanation (RDE) framework, which was introduced by \cite{RDE_original_2019} and later extended by \cite{in_distribution_cosmas_2020} by applying RDE to non-canonical input representations. Suppose $\Phi:\R^n\to \R^m$ is a pre-trained model, \eg, a classifier (with $m$ class labels) or a regression model (with $m$-dimensional output), where $n$ denotes the dimension of the model input. RDE produces an explanation for a model decision $\Phi(x)$ with $x\in\R^n$ as a relatively sparse mask $s\in\{0,1\}^n$ marking the relevant  input features in $x$. More precisely, RDE aims to solve the following constrained optimization problem over a mask $s\in\{0,1\}^n$:
\begin{align}\label{op: pixel rde}
    \min_{s\in \{0,1\}^n:\,\|s\|_0 \leq \ell} \;  \mathop{\mathbb{E}}_{v\sim \mathcal{V}} \Big[ d\Big(\Phi(x), \Phi(x\odot s + (1-s)\odot v)\Big)\Big] 
\end{align}
where $\odot$ denotes the Hadamard product (element-wise multiplication), $d(\Phi(x),\cdot)$ is a measure of distortion (\eg, $d(\Phi(x),\cdot)=\|\Phi(x) - \cdot \|_2$), $\calV$ is a distribution over input perturbations $v\in\R^n$, and $\ell\in\{1,...,n\}$ is a given sparsity level for the explanation mask $s$. A solution $s^*$ to the optimization problem (\ref{op: pixel rde}) masks relatively few components in the model input $x$ that suffice to approximately retain the model output $\Phi(x)$. This approach is in the spirit of rate-distortion theory, which deals with lossy compression of data. Therefore, \cite{RDE_original_2019} coined such explanations \emph{rate-distortion explanations} (RDEs). 

In practice, the RDE optimization problem is relaxed to continuous masks $s\in[0,1]^n$ solving:
\begin{align}
    \min_{s\in [0,1]^n} \mathop{\mathbb{E}}_{v\sim \mathcal{V}} \Big[ d\Big(\Phi(x), \Phi(x\odot s + (1-s)\odot v)\Big)\Big] +\lambda \norm{s}_1
\end{align}
In the relaxed optimization problem, the sparsity level of the mask is determined by $\lambda>0$ and an approximate solution can be found with stochastic gradient descent in $s\in[0,1]^n$ if $\Phi$ is differentiable. The authors of \cite{RDE_original_2019} applied the RDE method as described above to image classifiers in the pixel domain of images, where each mask entry $s_i\in[0,1]$ corresponds to the $i$-th pixel values. We refer to this method as \emph{Pixel RDE} throughout this work.

\section{RDE Reformulated and Reinterpreted}\label{section: rde reformulated}
Instead of applying RDE to the standard input representation $x=[x_1 \hdots x_n]^T$, we can apply RDE to a different representation of $x$ to answer a particular interpretation query. For example, consider a 1D-signal $x\in\R^n$: if we ask ``What is the smooth part in the signal $x$ that leads to the model decision $\Phi(x)$?", then we can apply RDE in the Fourier basis of $x$. Since frequency-sparse signals are smooth, applying RDE in the Fourier basis of $x$ extracts the relevant smooth part of the signal. To accommodate such interpretation queries, we reformulate RDE in Section \ref{subsection:general forumulation}. Finally, based on the reformulation, we reinterpret RDE in Section \ref{subsection:interpretation}. Later in Section \ref{section:cartoonRDE}, we use our reformulation and reinterpretation of RDE to derive and motivate CartoonX as a special case and novel explanation method tailored towards image classifiers.

\subsection{General Formulation}\label{subsection:general forumulation}

An input signal $x=[x_1,\hdots,x_n]^T$ is represented in a basis $\{b_1,\hdots,b_n\}$ as a linear combination
$\sum_{i=1}^n h_i b_i$ with coefficients $[h_i]_{i=1}^n$. As we argued above and demonstrate later on, some choices for a basis may be more suitable than others to explain a model decision $\Phi(x)$. Therefore, we define the RDE mask not only on the canonical input representation $[x_i]_{i=1}^n$ but also on a different representation $[h_i]_{i=1}^n$ with respect to a choice of basis $\{b_1,\hdots,b_n\}$. Examples of non-canonical choices for a basis include the Fourier basis and the wavelet basis. This work is centered around CartoonX, which applies RDE in the wavelet basis, \ie, a linear data representation. Nevertheless, there also exist other domains and interpretation queries where applying RDE to a non-linear data representation can make sense (see the interpretation query ``Is phase or magnitude more important for an audio classifier?" in \cite{in_distribution_cosmas_2020}). Therefore, we formulate RDE in terms of a data representation function $f:\prod_{i=1}^k\R^{c}\to\R^n,\; f(h_1, \hdots, h_k)=x$, which does not need to be linear and allows to mask $c$ channels in the input at once. In the important linear case and $c=1$, we have $f(h_1,\hdots,h_k)=\sum_{i=1}^k h_i b_i$, where $\{b_i,\hdots,b_k\}\subset\R^n$ are $k$ fixed vectors that constitute a basis. The case $c>1$ is useful when one wants to mask out several input channels at once, \eg, all color channels of an image, to reduce the number of entries in the mask that will operate on $[h_i]_{i=1}^k$. In the following, we introduce the important definitions of \emph{obfuscations}, \emph{expected distortion}, \emph{the RDE mask}, and \emph{RDE's $\ell_1$-relaxation}, which generalize the RDE framework of \cite{RDE_original_2019} to abstract input representations.

\subsubsection{Definitions}
The first two key concepts in RDE are \emph{obfuscations} and \emph{expected distortions}, which are defined below.
\begin{definition}[Obfuscations and expected distortions]\label{def: distortion and obsfustation}
Let $\Phi:\R^n\to\R^m$ be a model and $x\in \R^n $ a data point with a data representation $x =f(h_1,...,h_k)$ as discussed above. For every mask $s\in[0,1]^k$, let $\calV$ be a probability distribution over $\prod_{i=1}^k\R^{c}$. Then the \emph{obfuscation} of $x$ with respect to $s$ and $\mathcal{V}$ is defined as the random vector
$y \coloneqq f(s\odot h + (1-s)\odot v)$,
where $v\sim\mathcal{V}$, $(s\odot h)_i = s_i h_i\in\R^{c}$ and $((1-s)\odot v)_i= (1-s_i)v_i\in\R^{c}$, for $i\in\set{1, \hdots, k}$.
A choice for the distribution $\calV$ is called \emph{obfuscation strategy}.
Furthermore, the \emph{expected distortion} of $x$ with respect to the mask $s$ and the perturbation distribution $\mathcal{V}$ is defined as
$$
D(x,s,\mathcal{V}, \Phi)\coloneqq \mathop{\mathbb{E}}_{v\sim \mathcal{V}} \Big[ d\Big(\Phi(x), \Phi(y)\Big)\Big],
$$
where $d:\R^m\times \R^m\to\R_+$ is a measure of distortion between two model outputs.
\end{definition}
In the RDE framework, the explanation is given by a mask that minimizes distortion while remaining relatively sparse. The rate-distortion explanation mask is defined as follows.

\begin{definition}[The RDE mask]\label{def:rde mask}
In the setting of Definition \ref{def: distortion and obsfustation}, we define the \emph{RDE mask} as a solution $s^*(\ell)$ to the minimization problem
\begin{align}\label{eq:rde mask}
    \min_{s\in \{0,1\}^k} \quad D(x,s,\mathcal{V}, \Phi) \quad \text{ s.t. } \quad \norm{s}_0 \leq \ell,
\end{align}
where $\ell\in \set{1, \hdots, k}$ is the desired level of sparsity.
\end{definition}

\begin{figure}[h]
     \centering
     \begin{subfigure}[b]{0.45\textwidth}
    \centering
    \includegraphics[width=\linewidth]{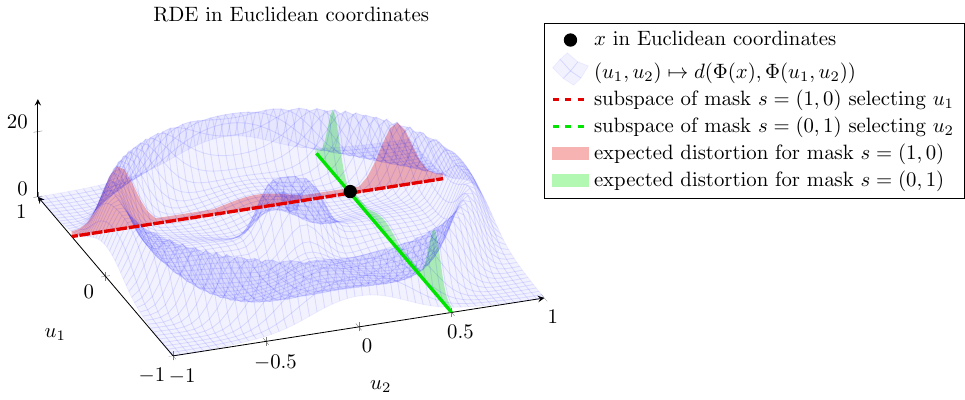}
    \caption{}
    \label{fig:explaining misclassfications}
     \end{subfigure}
     \hfill
     \begin{subfigure}[b]{0.45\textwidth}
    \centering
    \includegraphics[width=\linewidth]{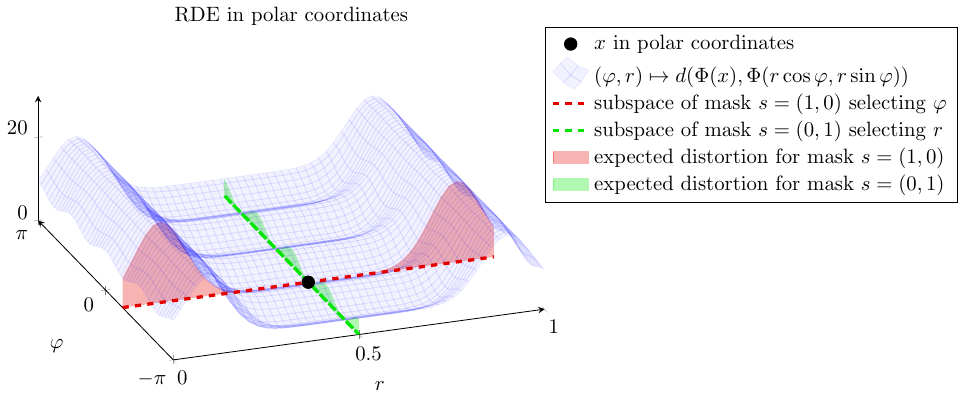}
    \caption{}
    \label{fig:comparison snails}
     \end{subfigure}
    \caption{RDE for a hypothetical toy-example in (a) Euclidean coordinates and (b) polar coordinates. Here, the RDE mask can find low expected distortion in polar coordinates but not in Euclidean coordinates. Therefore, in this example, polar coordinates are more appropriate to explain $\Phi(x)$, and RDE would determine that the angle $\varphi$, not the magnitude $r$, is relevant for $\Phi(x)$.}
    \label{fig: geometric view rde}
\end{figure}

Geometrically, the RDE mask $s$ is associated with a particular subspace. The complement mask $(1-s)$ can be seen as selecting a large stable subspace of $\Phi$, where each point represents a possible perturbation in unselected coefficients in $h$. The RDE mask minimizes the expected distortion along its associated subspace, which requires non-local information of $\Phi$. We illustrate this geometric view of RDE in Figure \ref{fig: geometric view rde} with a toy example for a hypothetical classifier $\Phi: \R^2\to\R^m$ and two distinct input representations: (1) Euclidean coordinates, \ie, $f$ is the identity in $x=f(h)$, and (2) polar coordinates, \ie, $f(h)=(h_2\cos h_1, h_2\sin h_1)=x$. 
In the example, we assume $\mathcal{V}$ to be a uniform distribution on $[-1,1]^2$ in the Euclidean representation and a uniform distribution on $[-\pi,\pi]\times[0,1]$ in the polar representation. The expected distortion associated with the masks $s=(1,0)$ and $s=(0,1)$ is given by the red and green shaded area, respectively. The RDE mask aims for low expected distortion, and hence, in polar coordinates, the RDE mask would be the green subspace, \ie, $s=(0,1)$. On the other hand, in Euclidean coordinates, neither $s=(1,0)$ nor $s=(0,1)$ produces a particularly low expected distortion, making the Euclidean explanation less meaningful than the polar explanation. 
The example illustrates why certain input representations can yield more meaningful explanatory insight for a given classifier than others---an insight that underpins our novel CartoonX method. 
Moreover, the plot in polar coordinates illustrates why the RDE mask cannot be simply chosen with local distortion information, \eg, with the lowest eigenvalue of the Hessian of $h \mapsto d(\Phi(x), \Phi(f(h)))$: the lowest eigenvalue  in polar coordinates belongs to the red subspace and does not see the large distortion on the tails.

As was shown by \cite{RDE_original_2019}, the RDE mask from Definition \ref{def:rde mask} cannot be computed efficiently for non-trivial input sizes. Nevertheless, one can find an approximate solution by considering continuous masks $s\in[0,1]^k$ and encouraging sparsity through the $\ell_1$-norm.

\begin{definition}[RDE's $\ell_1$-relaxation ]\label{def:ell_1 relaxation}
In the setting of Definition \ref{def: distortion and obsfustation}, we define \emph{RDE's $\ell_1$-relaxation} as a solution $s^*(\lambda)$ to the minimization problem
\begin{align}\label{eq:relax_problem_ell1}
    \min_{s\in [0,1]^k} \quad D(x,s,\mathcal{V}, \Phi) + \lambda \|s\|_1,
\end{align}
where $\lambda>0$ is a hyperparameter for the sparsity level. 
\end{definition}
The $\ell_1$-relaxation above can be solved with stochastic gradient descent (SGD) over the mask $s$ while approximating $D(x,s,\mathcal{V}, \Phi)$ with i.i.d. samples from $v\sim \calV$.

\subsubsection{Obfuscation Strategies}\label{subsubsec: obfuscation strategies}
An obfuscation strategy is defined by the choice of the perturbation distribution $\calV$. Common choices are Gaussian noise \cite{RDE_original_2019,Fong_vedaldi_2017}, blurring \cite{Fong_vedaldi_2017}, constants \cite{Fong_vedaldi_2017}, and inpainting GANs \cite{in_distribution_cosmas_2020,chang_GAN_explanation_2018}. Inpainting GANs train a generator $G(s,z,h)$ ($z$ denotes random latent factors) such that for samples $v\sim G(s,z,h)$ the obfuscation $f(s\odot h + (1-s)\odot v)$
remains in the data manifold. In our work, we refrain from using an inpainting GAN due to the following reason: it is hard to tell whether a GAN-based mask did not select coefficients because they are unimportant or because the GAN can easily inpaint them from a biased context (\eg, a GAN that always inpaints a car when the mask shows a traffic light). We want to explain a black-box method transparently, which is why we opt for a simple distribution on the price of not accurately representing the data distribution. We choose a simple and well-understood obfuscation strategy, which we call \emph{Gaussian adaptive noise}. It works as follows: Let $A_1,..., A_j$ be a pre-defined choice for a partition of $\{1,\hdots,k\}$. For $i=1,...,j$, we compute the empirical mean and empirical standard deviation for each $A_i$:
\begin{align}
    &\mu_i\coloneqq \frac{\sum_{a\in A_i, t=1,...,d_a} h_{at}}{\sum_{a\in A_i}d_a} ,\;
    &\sigma_i\coloneqq \sqrt{\frac{1}{\sum_{a\in A_i}d_a}\sum_{a\in A_i,t=1,...,d_a}(\mu_i - h_{at})^2}
\end{align}
The adaptive Gaussian noise strategy then samples $v_{at}\sim \mathcal{N}(\mu_i, \sigma_i^2)$ for all members $a\in A_i$ and channels $t=1,...,d_a$. We write $v\sim \mathcal{N}(\mu,\sigma^2)$ for the resulting Gaussian random vector $v\in \prod_{i=1}^k\R^{c}$. For Pixel RDE, we  only use one set $A_1=\{1,...,k\}$ for all $k$ pixels. In CartoonX, which represents input signals in the discrete wavelet domain, we partition $\{1,...,k\}$ along the scales of the discrete wavelet transform.

\subsubsection{Measures of distortion}\label{subsubsec: measures of distortion}
There are various choices for the measure of distortion $d(\Phi(x),\Phi(y))$. For example, one can take the squared distance in the post-softmax probability of the predicted label for $x$, \ie, 
$d\big(\Phi(x),\Phi(y) \big) \coloneqq \big(\Phi_{j^*}(x)- \Phi_{j^*}(y) \big)^2$,
where $j^*\coloneqq \argmax_{i=1,...,m}\Phi_i(x)$ and $\Phi(x)$ is assumed to be the post-softmax probabilities of a neural net. Alternatively, one could also choose $d(\Phi(x),\Phi(y))$ as the $\ell_2$-distance or the KL-Divergence in the post-softmax layer of $\Phi$. In our experiments for CartoonX, we found that these choices had no significant effect on the explanation (see Figure \ref{fig:sensitivity distortion}).

\subsection{Interpretation}\label{subsection:interpretation}
The philosophy of the generalized RDE framework is that an explanation for a decision $\Phi(x)$ on a generic input signal $x=f(h)$ should be some simplified version of the signal, which is interpretable to humans. The simplification is achieved by demanding sparsity in a suitable representation system $h$, \emph{which sparsely represents the class of explanations that are desirable for the interpretation query}. This philosophy is the fundamental premise of CartoonX, which aims to answer the interpretation query \emph{``What is the relevant piece-wise smooth part of the image for a given image classifier?"}. CartoonX first employs RDE on a representation system $x=f(h)$ that sparsely represents piece-wise smooth images and finally visualizes the relevant piece-wise smooth part as an image back in pixel space. In the following section, we explain why wavelets provide a suitable representation system in CartoonX, discuss the CartoonX implementation, and  evaluate CartoonX qualitatively and quantitatively on  ImageNet.

\section{CartoonX}\label{section:cartoonRDE}
The focus of this paper is \emph{CartoonX}, a novel explanation method---tailored to image classifications---that we obtain as a special case of our generalized RDE framework formulated in Section \ref{section: rde reformulated}.  CartoonX first performs RDE in the discrete wavelet position-scale domain of an image $x$, and finally, visualizes the wavelet mask $s$ as a piece-wise smooth image in pixel space.
Wavelets provide optimal representations for piece-wise smooth 1D functions \cite{devore_1998}, and represent 2D piece-wise smooth images, also called \emph{cartoon-like images} \cite{kutyniok2010}, efficiently as well \cite{Romberg06wavelet-domainapproximation}. In particular, sparse vectors in the wavelet coefficient space encode cartoon-like images reasonably well \cite{STEPHANE2009535}---certainly better than sparse pixel representations. Moreover, wavelets constitute an established tool in image processing \cite{mallatbook}.

\begin{figure}[h]
    \centering
    \hspace*{-0.3cm}
    \includegraphics[scale=0.62]{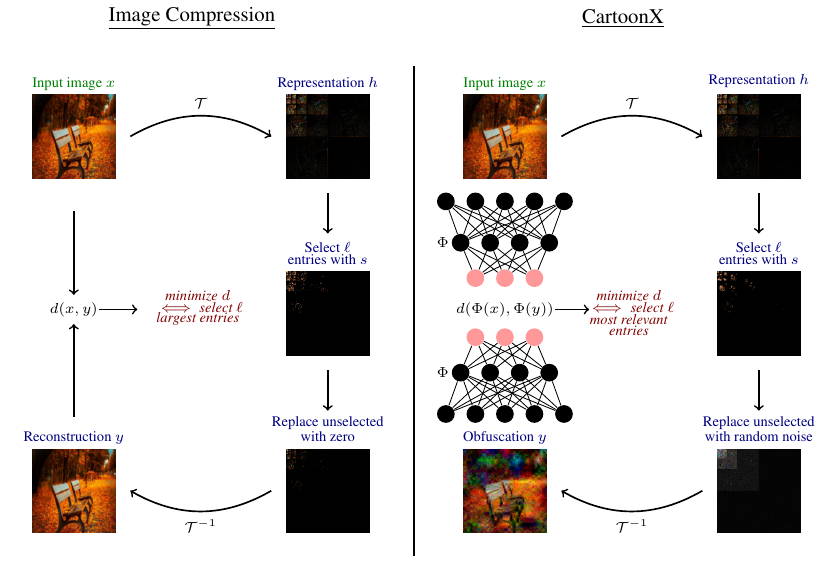}
    \caption{CartoonX shares many interesting parallels to wavelet-based image compression. Distortion is denoted as $d$, $\Phi$ is an image classifier, $h$ denotes the discrete wavelet coefficients, $\mathcal{T}$ is the discrete wavelet transform, and $\ell$ is the coefficient budget. }
    \label{fig: diagram RDE}
\end{figure}

The optimization process underlying CartoonX produces sparse vectors in the discrete wavelet coefficient space, which results in cartoon-like images as explanations. This is the fundamental difference to Pixel  RDE, which produces rough, jittery, and pixel-sparse explanations. Cartoon-like images provide a natural model of simplified images. Since the goal of the RDE framework is to generate an easy to interpret simplified version of the input signal, we argue that CartoonX explanations are more appropriate for image classification than Pixel RDEs. Previous work, such as Grad-CAM \cite{Grad_CAM_2017}, produces smooth explanations, which also avoid jittery explanations. CartoonX produces \emph{roughly piece-wise smooth explanations and not smooth explanations}, which we believe to be more appropriate for images, since smooth explanations cannot preserve edges well. Moreover, we believe that CartoonX enforces piece-wise smoothness in a mathematically more natural manner than explicit smoothness regularization (as in \cite{Fong_2019_ICCV}) because wavelets sparsely represent piece-wise smooth signals well. Therefore, CartoonX does not rely on additional smoothness hyperparameters.

CartoonX exhibits interesting parallels to wavelet-based image compression. In image compression, distortion is minimized in the image domain, which is equivalent to selecting the $\ell$ largest entries in the discrete wavelet transform (DWT) coefficients. CartoonX minimizes distortion in the model output of $\Phi$, which translates to selecting the $\ell$ most relevant entries in the DWT coefficients.
The objective in image compression is efficient data representation, \ie, producing minimal data distortion with a budget of $\ell$ entries in the DWT coefficients. Conversely, in CartoonX, the objective is extracting the relevant piece-wise smooth part, \ie, producing minimal model distortion with a budget of $\ell$ entries in the DWT coefficients. 
We illustrate this connection in Figure \ref{fig: diagram RDE}---highlighting once more the \emph{rate-distortion} spirit of the RDE framework.

\begin{figure}[h]
    \centering
    \includegraphics[scale=0.6]{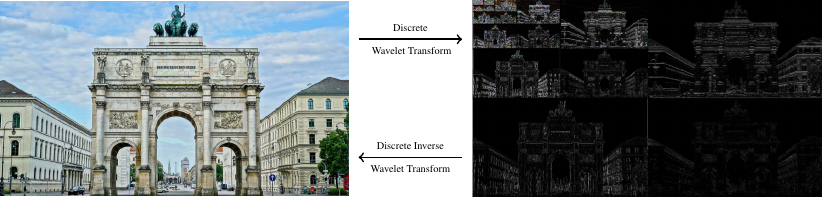}
	\caption{Visualization of the DWT coefficients for five scales. Three L-shaped sub-images describe coefficients for details in vertical, horizontal, and diagonal orientation at a particular scale. The largest sub-images (the outer L-shape) belong to the lowest scale, \ie, the highest resolution. The smaller L-shaped sub-images gradually build up to higher scales, \ie, lower resolution features. }
	\label{fig:dwt example}
\end{figure}

\subsection{Implementation}
An image $x\in[0,1]^n$ with $c\in\{1,3\}$ channels, $k\in\N$ pixels can be represented in a wavelet basis by computing its DWT, which is defined by the number of scales $J \in \set{{1, \hdots, \lfloor \log_2 k \rfloor}}$,  the padding mode, and a choice of the mother wavelet (\eg, Haar or Daubechies). For images, the DWT computes four types of coefficients: details in (1) horizontal, (2) vertical, and (3) diagonal orientation at scale $j\in\set{{1, \hdots, J}}$, and  (4) coefficients of the image at the very coarsest resolution. We briefly illustrate the DWT for an example image in Figure \ref{fig:dwt example}.

CartoonX, as described in Algorithm \ref{alg: cartoon RDE}, computes the RDE mask in the wavelet domain of images. More precisely, for the data representation $x = f(h)$,
we choose $h$ as the concatenation of all the DWT coefficients along the channels, \ie, $h_i\in\R^{c}$. The representation function $f$ is then the discrete inverse wavelet transform, \ie, the summation of the DWT coefficients times the DWT basis vectors. We optimize the mask $s\in[0,1]^k$ on the DWT coefficients $[h_1,\hdots,h_k]^T$ to minimize RDE's $\ell_1$-relaxation from Definition \ref{def:ell_1 relaxation}. For the obfuscation strategy $\mathcal{V}$, we use adaptive Gaussian noise with a partition by the DWT scale (see Section \ref{subsubsec: obfuscation strategies}), \ie, we compute the empirical mean and standard deviation per scale. 
To visualize the final DWT mask $s$ as a piece-wise smooth image in pixel space, we multiply the mask with the DWT coefficients of the greyscale image $\hat x$ of $x$ before inverting the product back to pixel space with the inverse DWT. The pixel values of the inversion are finally clipped into $[0,1]$ as are obfuscations during the RDE optimization to avoid overflow (we assume here the pixel values in $x$ are normalized into $[0,1]$). The clipped inversion in pixel space is the final CartoonX explanation.

\RestyleAlgo{ruled} 
\SetKwInput{kwHparams}{Hyperparameters}
\SetKwInput{kwInit}{Initialization}

\begin{algorithm}[hbt!]

\caption{CartoonX}\label{alg: cartoon RDE}
\KwData{Image $x\in[0,1]^n$ with $c$ channels and $k$ pixels, pre-trained classifier $\Phi$.}
 \kwInit{Initialize mask $s\coloneqq[1,...,1]^T$ on\\ \\ DWT coefficients $h=[h_1,...,h_k]^T$ with $x=f(h)$, where $f$ is the inverse DWT. Choose sparsity level $\lambda>0$, number of steps $N$,  number of noise samples $L$, and measure of distortion $d$.}
  \For{$i\gets1$ \KwTo $N$}{
    Sample $L$ adaptive Gaussian noise samples $v^{(1)},...,v^{(L)}\sim \mathcal{N}(\mu,\sigma^2)$\;
    Compute obfuscations $y^{(1)},..., y^{(L)}$ with $y^{(i)}\coloneqq f(h\odot s + (1-s)\odot v^{(i)})$\;
    Clip obfuscations into $[0,1]^{n}$\;
    Approximate expected distortion $\hat D(x,s,\Phi)\coloneqq \sum_{i=1}^Ld(\Phi(x),\Phi(y^{(i)}))^2/L$\;
    Compute loss for the mask, \ie, $\ell(s)\coloneqq \hat D(x,s,\Phi) + \lambda \|s\|_1$\;
    Update mask $s$ with gradient descent step using $\nabla_s \ell(s)$ and clip $s$ back to $[0,1]^{k}$\;
    }
    Get DWT coefficients $\hat h$ for greyscale image $\hat x$ of $x$\;
    Set ${\mathcal{E}}\coloneqq f(\hat h \odot s)$ and finally clip ${\mathcal{E}}$ into $[0,1]^{k}$\;
\end{algorithm}

\subsection{Experiments}
We compare CartoonX to the closely related Pixel RDE \cite{RDE_original_2019} and several other state-of-the-art explanation methods, \ie,
Integrated Gradients \cite{Integrated_gradient_2017_sundararajan}, Smoothgrad \cite{Smooth_Grad_2017}, Guided Backprop \cite{guided_backprop_2015}, LRP \cite{Layerwise_relevance_prop2015}, Guided Grad-CAM \cite{Selvaraju2019GradCAMVE}, Grad-CAM \cite{Selvaraju2019GradCAMVE}, and LIME \cite{LIME_2016}. Our experiments use the pre-trained ImageNet classifiers MobileNetV3-Small \cite{Howard2019SearchingFM} (67.668\% top-1 acc.) and VGG16 \cite{brusilovsky:simonyan2014very}  (71.592\% top-1 acc.). Images were preprocessed to have $256\times256$ pixel values in $[0,1]$.  
Throughout our experiments with CartoonX and Pixel RDE, we used the Adam optimizer \cite{adam}, a learning rate of $\epsilon=0.001$, $L=64$  adaptive Gaussian noise samples, and $N=2000$ steps. Several different sparsity levels were used. We specify the sparsity level in terms of the number of mask entries $k$, \ie, by choosing the product $\lambda k$. Pixel RDE typically requires a smaller sparsity level than CartoonX. We chose $\lambda k\in[20,80]$ for CartoonX and $\lambda k\in[3,20]$ for Pixel RDE. The obfuscation strategy for Pixel RDE was chosen as Gaussian adaptive noise with mean and standard deviation computed for all pixel values (see Section \ref{subsubsec: obfuscation strategies}). We implemented the DWT for CartoonX with the Pytorch Wavelets package, which is compatible with PyTorch gradient computations, and chose the Daubechies 3 wavelet system with $J=5$ scales and zero-padding. For the Integrated Gradients method, we used $100$ steps, and for the Smoothgrad method, we used $10$ samples and a standard deviation of $0.1$.

\subsubsection{Interpreting CartoonX}\label{sec: interpreting}
In order to correctly interpret CartoonX, we briefly review important properties of the DWT. To cover a large area in an image with a constant value or slowly and smoothly changing gray levels, it suffices to select very few high-scale wavelet coefficients.  Hence, for the wavelet mask in CartoonX, it is cheap to cover large image regions with constant or blurry values. Conversely, one needs many high-scale wavelet coefficients to produce fine details such as edges in an image, so fine details are expensive for CartoonX. Hence, the fine details present in the CartoonX are important features for the outcome of the classifier, and fine image features that are replaced by smooth areas in CartoonX are not important for the classifier.  
\begin{figure}[h]
    \centering
    \vspace*{-.2cm}
    \hspace*{.2cm}
    \includegraphics[scale=.8]{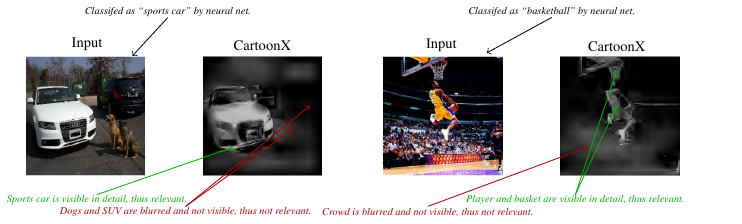}
        \caption{The CartoonX explanation is an image that suffices to retain the classification decision. For the sports car, CartoonX blurs out the SUV and the dogs. This means the dog and the SUV are irrelevant. For the basketball, the crowd is blurred out. This means the crowd is not relevant since the player and the basket with the crowd blurred out retains the classification as ``basketball". The left example also shows that CartoonX is class-discriminative since it blurs out the dogs and the SUV, which belong to other classes.}
    \label{fig:how to interpret}
\end{figure}
It is important to keep in mind that the final CartoonX explanation is a visualization of the wavelet mask in pixel space, and \emph{should not be interpreted as a pixel-mask or ordinal pixel-attribution.} CartoonX is not a saliency-map or heatmap but an explanation that is to be interpreted as an image that suffices
to retain the classification decision. 
We illustrate this point in Figure \ref{fig:how to interpret} with two examples.


\subsubsection{Qualitative Evaluation}\label{sec: qualitative}

\begin{figure}[h]
     \centering
     \begin{subfigure}[b]{0.34\textwidth}
    \centering
    \hspace*{-.3cm}
    \includegraphics[width=1.15\linewidth]{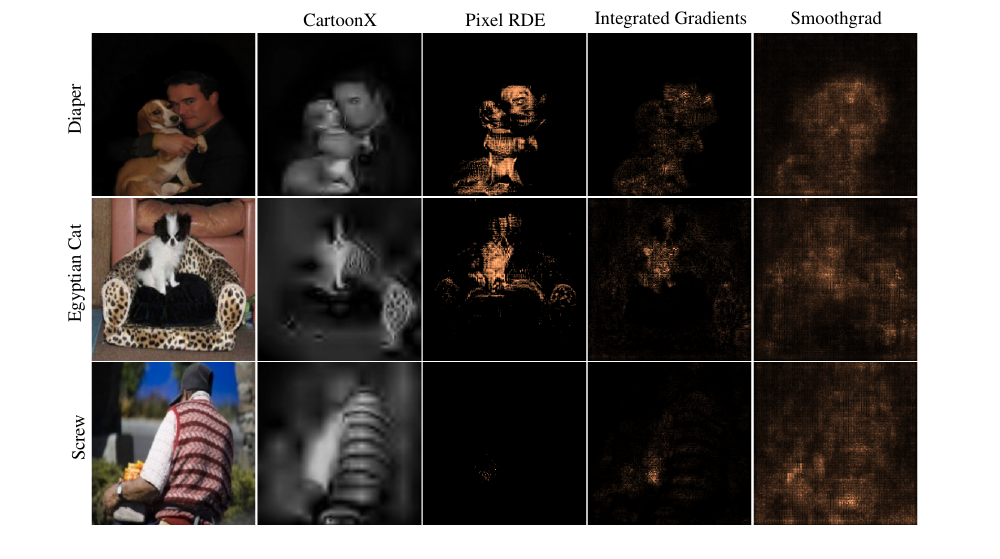}
    \caption{}
    \label{fig:explaining misclassfications}
     \end{subfigure}
     \hfill
     \begin{subfigure}[b]{0.6\textwidth}
    \centering
    \includegraphics[width=\linewidth]{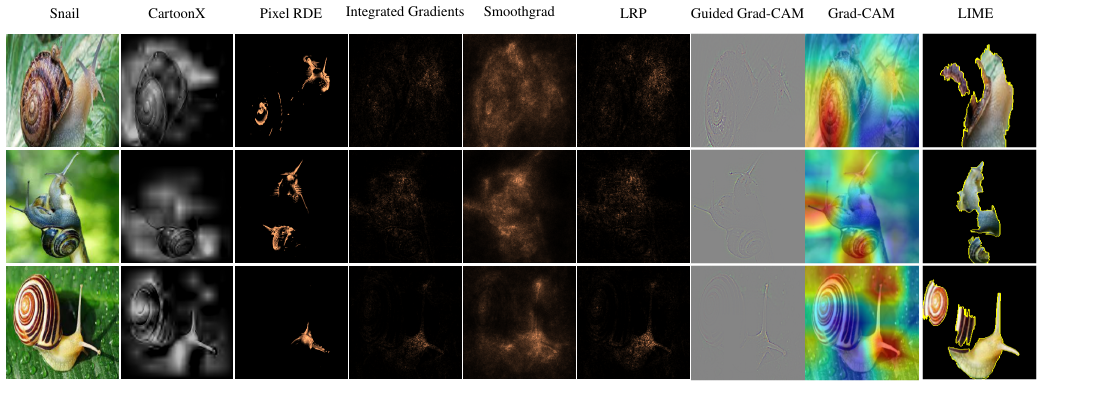}
    \caption{}
    \label{fig:comparison snails}
     \end{subfigure}
        \caption{(a) Each row compares CartoonX explanations of misclassifications by MobileNetV3-Small. The predicted label is depicted next to each misclassified image. (b) Comparing CartoonX explanations for VGG16 for three different images of correctly classified snails.}
        \label{fig:three graphs}
\end{figure}

In practice, explaining misclassifications is particularly relevant since good explanations can pinpoint model biases and causes for model failures. 
In Figure \ref{fig:explaining misclassfications}, we illustrate how CartoonX can help explain misclassified examples by revealing classifier-relevant piece-wise smooth patterns that are not easily visible in other pixel-based methods. In the first row in Figure \ref{fig:explaining misclassfications}, the input image shows a man holding a dog that was classified as a ``diaper". CartoonX shows the man not holding a dog but a baby, possibly revealing that the neural net associated diapers with babies and babies with the pose with which the man is holding the dog. In the second row, the input image shows a dog sitting on a chair with leopard patterns. The image was classified as an ``Egyptian Cat", which can exhibit leopard-like patterns. CartoonX exposes the Egyptian cat by connecting the dog’s head to parts of the armchair forming a cat’s torso and legs. In the last row, the input image displays the backside of a man wearing a striped sweater that was classified as a ``screw". CartoonX reveals how the stripe patterns look like a screw to the neural net.

\begin{figure}[h]
     \centering
     \begin{subfigure}[b]{0.55\textwidth}
    \centering
    \includegraphics[width=\linewidth]{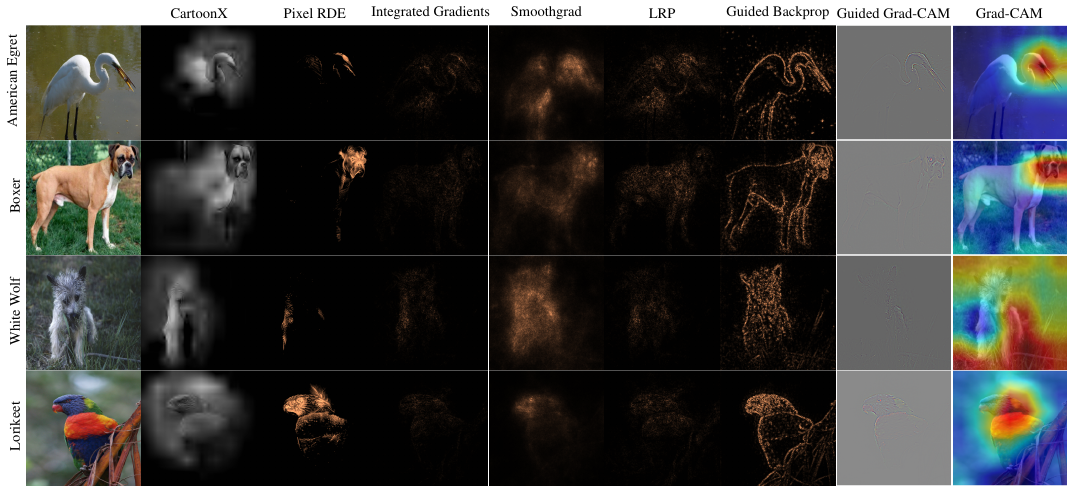}
    \caption{CartoonX on random samples.}
    \label{fig:random evaluation}
     \end{subfigure}
     \hfill
     \begin{subfigure}[b]{0.4\textwidth}
    \centering
    \includegraphics[width=.88\linewidth]{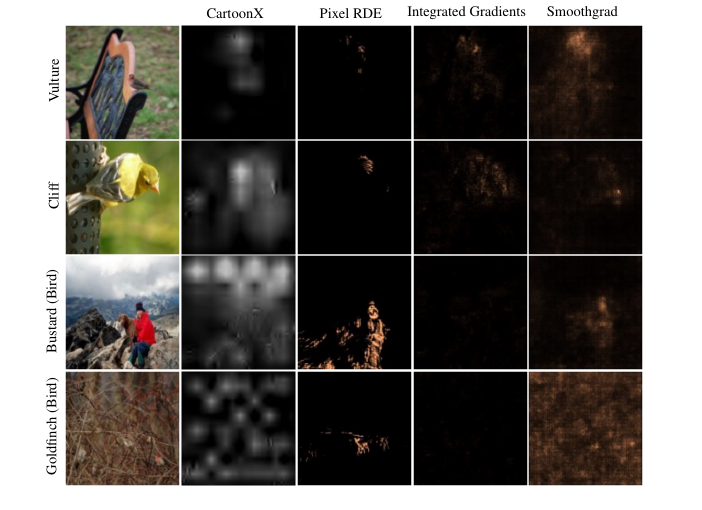}
    \caption{Examples of CartoonX failures.}
    \label{fig:failures}
     \end{subfigure}
        
        \caption{On random Imagenet samples, CartoonX consistently produces interpretable explanations. Established explanation methods tend to also be difficult to interpret on CartoonX's failure examples.}
        \label{fig:three graphs}
\end{figure}

Figure \ref{fig:comparison snails} further compares CartoonX explanations of correct classifications by VGG16. We also compare CartoonX on random ImageNet samples in Figure \ref{fig:random evaluation} to provide maximal transparency and fair qualitative comparison. In Figure \ref{fig:failures}, we also show failures of CartoonX. These are examples of explanations that are not interpretable and seem to fail at explaining the model prediction. Notably, most failure examples are also not particularly well explained by other state-of-the-art methods. It is challenging to state with certainty the underlying reason for the CartoonX failures since there it is always possible that the neural net bases its decision on non-interpretable grounds.

\vspace{-.5cm}

\subsubsection{Quantitative Evaluation}

To compare CartoonX quantitatively against other explanation methods, we computed explanations for 100 random ImageNet samples and ordered the image coefficients (for CartoonX the wavelet coefficients) by their respective relevance score. 
\begin{figure}[h]
     \centering
     \begin{subfigure}[b]{0.3\textwidth}
         \centering
         \includegraphics[width=.9\textwidth]{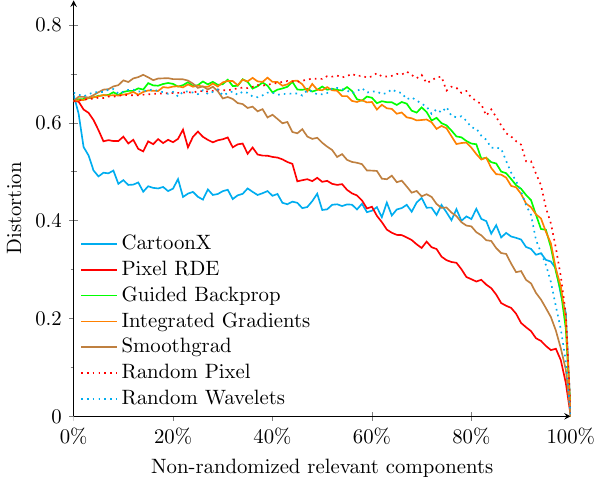}
         \caption{}
         \label{fig:non-randomized}
     \end{subfigure}
     \hfill
     \begin{subfigure}[b]{0.3\textwidth}
         \centering
         \includegraphics[width=.9\textwidth]{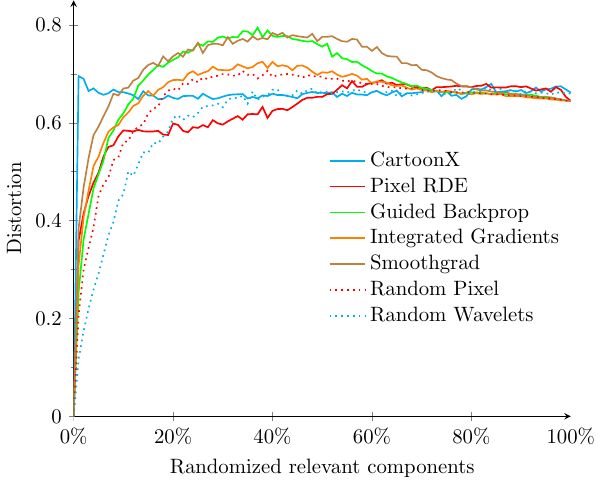}
         \caption{}
         \label{fig:randomized}
     \end{subfigure}
     \hfill
     \begin{subfigure}[b]{0.3\textwidth}
         \centering
         \includegraphics[width=.8\textwidth]{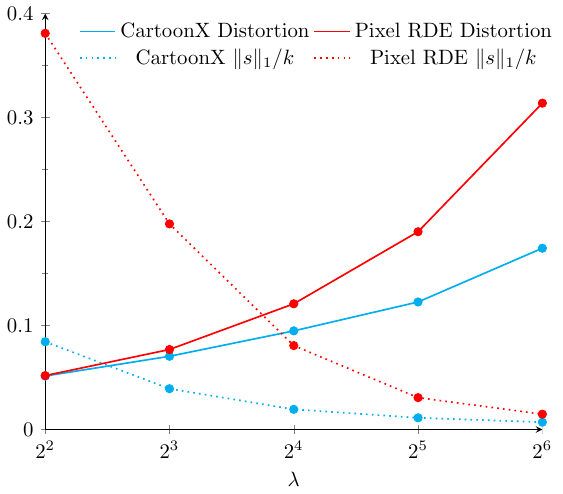}
         \caption{}
         \label{fig:lambda_curve}
     \end{subfigure}
        
        \caption{ In (a) the best explanation exhibits steepest early decay. In (b) best explanation exhibits sharpest early increase. In (c) best explanation exhibits lowest distortion and lowest normalized $\ell_1$-norm of mask (\ie, highest sparsity).}
        \label{fig:three graphs}
\end{figure}
Figure \ref{fig:non-randomized} plots the rate-distortion curve, \ie, the distortion achieved in the model output (measured as the $\ell_2$-norm in the post-softmax layer) when keeping the most relevant coefficients 
and randomizing the others. We expect a good explanation to have the most rapid decaying rate-distortion curve for low rates (non-randomized components), which is the case for CartoonX. Note that the random baseline in the wavelet representation is not inherently more efficient than the random baseline in the pixel representation. 
Moreover, Figure \ref{fig:randomized} plots the achieved distortion versus the fraction of randomized relevant components. Here, we expect a good explanation to have the sharpest early increase, which CartoonX again realizes.
Lastly, Figure \ref{fig:lambda_curve} plots the distortion and non-sparsity (measured as the normalized $\ell_1$-norm) of the RDE mask  for Pixel RDE and CartoonX at different $\lambda$ values. The plot underscores the efficiency advantage of CartoonX over Pixel RDE since CartoonX achieves lower distortion and higher sparsity throughout all $\lambda$ values. For all three plots,  random perturbations were drawn from the adaptive Gaussian distribution described in Section \ref{subsubsec: obfuscation strategies}.

\subsubsection{Sensitivity to Hyperparameters}\label{sec: sensitivity}

\begin{figure}
     \centering
\begin{subfigure}[h]{0.5\linewidth}
    \centering
    \hspace*{-1.7cm}
    \includegraphics[width=\linewidth]{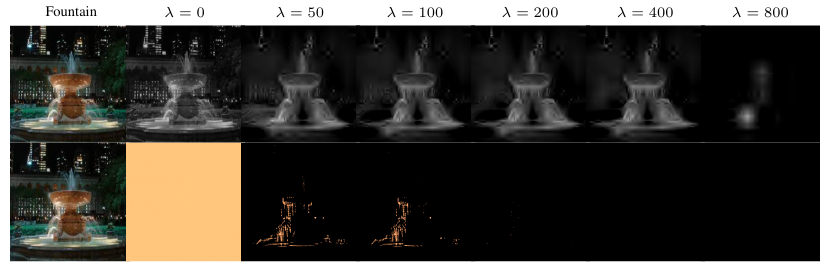}
    \caption{}
    \label{fig:sensitivity lambda}
\end{subfigure}
\begin{subfigure}[h]{.3\linewidth}
\centering
    \hspace*{-0.3cm}
    \includegraphics[width=\linewidth]{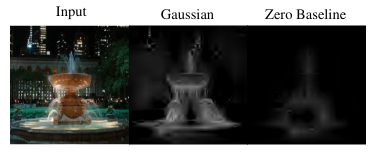}
    \caption{}
    \label{fig:sensitivity distritbution}
\end{subfigure}
\begin{subfigure}[h]{0.45\linewidth}
\centering
\hspace*{-0.8cm}
\vspace*{-0.16cm}
    \includegraphics[width=1.2\linewidth]{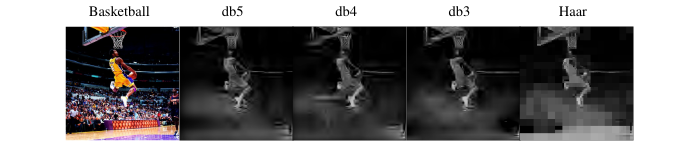}
    \caption{}
    \label{fig:sensitivity motherwavelet}
\end{subfigure}
\begin{subfigure}[h]{0.45\linewidth}
   \centering
    \includegraphics[width=\linewidth]{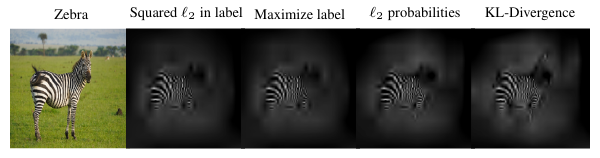}
    \caption{}
    \label{fig:sensitivity distortion}
\end{subfigure}
\caption{(a) Top row depicts CartoonX, and the bottom row depicts Pixel RDE, for increasing values of $\lambda$. CartoonX for different (b) perturbation distributions, (c) mother wavelets, (d) distortion measures.}

\end{figure}

We compare qualitatively CartoonX's sensitivity to its primary hyperparameters.
 Figure \ref{fig:sensitivity lambda} plots CartoonX explanations and Pixel RDEs for increasing $\lambda$. We conistently find that CartoonX is less sensitive than Pixel RDE to $\lambda$. In practice, this means one can find a suitable $\lambda$ faster for CartoonX than for Pixel RDE.
 Note that for $\lambda=0$, Pixel RDE is entirely yellow because the mask is initialized as $s=[1\hdots 1]^T$ and $\lambda=0$ provides no incentive to make $s$ sparser. For the same reason, CartoonX is simply the greyscale image when $\lambda=0$.
 Figure \ref{fig:sensitivity distritbution} plots CartoonX explanations for two choices of $\calV$: (1)  Gaussian adaptive noise (see Section \ref{subsubsec: obfuscation strategies}) and (2) constant zero perturbations.
We observe that the Gaussian adaptive noise gives much more meaningful explanations than the simple zero baseline perturbations. 
Figure \ref{fig:sensitivity distortion} plots CartoonX explanations for four choices of $d(\Phi(x), \Phi(y))$, where $x$ is the original input, $y$  is the RDE obfuscation, and $\Phi$ outputs post-softmax probabilities: (1) squared $\ell_2$ in probability of predicted label $j^*$, (2) $d(\Phi(x),\cdot)=\|\Phi_{j^*}(x)-1\|$, \ie, distance that maximizes probability of predicted label, (3) $\ell_2$ in post-softmax, (4) KL-Divergence in post-softmax.
We do not observe a significant effect by the distortion measure on the explanation.
Finally, in Figure \ref{fig:sensitivity motherwavelet} we compare the effect of the mother wavelet in the DWT on the CartoonX explanation. All choices of mother wavelets (labeled as in the Pytorch Wavelets package) provide consistent explanations except for the Haar wavelet, which produces images built of large square pixels.

\subsubsection{Limitations}

For MobileNetV3-Small, an image of $256\times256$ pixels, 16 noise samples, and 2000 optimization steps, we reported a runtime of 45.10s for CartoonX and 34.09s for Pixel RDE on the NVIDIA Titan RTX GPU. CartoonX is only slightly slower than Pixel RDE. However, like other perturbation-based methods, CartoonX is significantly slower than gradient or propagation-based methods, which only compute a single or few forward and backward passes and are very fast (Integrated Gradients computes an explanation in 0.48s for the same image, model, and hardware). We acknowledge that the runtime for CartoonX in its current form constitutes a considerable limitation for many critical applications. However, we are confident that we can significantly reduce the runtime in future work by either learning a strong initial wavelet mask with a neural net or even learning the final wavelet mask with a neural net, similar to the real time image saliency work in \cite{Dabkowski2017RealTI}. Finally, solving RDE's $\ell_1$-relaxation requires access to the model's gradients. Hence, CartoonX is limited to differentiable models.

\section{Conclusion}
CartoonX is the first explanation method for differentiable image classifiers based on wavelets.
We corroborated experimentally that CartoonX can reveal novel explanatory insight and achieves a better rate-distortion than  state-of-the-art methods. Nonetheless, CartoonX is still computationally quite expensive, like other perturbation-based explanation methods. In the future, we hope to devise new techniques to speed up the runtime for CartoonX and study the effect of using inpainting GANs for perturbations. We believe CartoonX is a valuable new explanation method for practitioners and potentially a great source of inspiration for future explanation methods tailored to specific data domains. 

\subsubsection{Acknowledgments}
GK was supported in part by the ONE Munich Strategy Forum as well as by Grant DFG-SFB/TR 109, Project C09 and DFG-SPP-2298, KU 1446/31-1 and KU 1446/32-1.

%
%
\bibliographystyle{splncs04}
\bibliography{egbib}
\end{document}